\documentclass[10pt,twocolumn,letterpaper]{article}

\usepackage[pagenumbers]{cvpr2024/cvpr} 

\usepackage[dvipsnames]{xcolor}

\usepackage{subfiles}
\usepackage{term_to_use}
\usepackage{boldline}
\usepackage{enumitem}
\usepackage{multirow}
\usepackage{graphicx}
\usepackage{amsmath}
\usepackage{amssymb}
\usepackage{caption}

\definecolor{cvprblue}{rgb}{0.21,0.49,0.74}
\usepackage[pagebackref,breaklinks,colorlinks,citecolor=cvprblue]{hyperref}

\def\paperID{17554} 
\def\confName{CVPR}
\def\confYear{2024}

\title{WIDIn: Wording Image for Domain-Invariant Representation \\ in Single-Source Domain Generalization}

\author{Jiawei Ma \quad Yulei Niu \quad Shiyuan Huang \quad Guangxing Han \quad Shih-Fu Chang\\
Columbia University\\
{\tt\small \{jiawei.m, yulei.niu\}@columbia.edu}}

\begin{document}
\maketitle

\begin{abstract}
Language has been useful in extending the vision encoder to data from diverse distributions without empirical discovery in training domains. However, as the image description is mostly at coarse-grained level and ignores visual details, the resulted embeddings are still ineffective in overcoming complexity of domains at inference time.
We present a self-supervision framework WIDIn, \underline{W}ording \underline{I}mages for \underline{D}omain-\underline{In}variant representation, to disentangle discriminative visual representation, by only leveraging data in a single domain and without any test prior. 
Specifically, for each image, we first estimate the language embedding with fine-grained alignment, which can be consequently used to adaptively identify and then remove domain-specific counterpart from the raw visual embedding. 
WIDIn can be applied to both pretrained vision-language models like CLIP, and separately trained uni-modal models like MoCo and BERT. Experimental studies on three domain generalization datasets demonstrate the effectiveness of our approach.

\end{abstract}

\section{Introduction}

Transferring models to the data drawn from different domains is important for real-world applications.
For example, given a model pre-trained to recognize airplanes while the training data are only in natural environments in the daytime (\ie, single domain), it should also detect airplanes in cartoon drawing and night vision (\ie, target domains unseen in training). 
However, since the model may utilize the information that is unique to the training domain for task performing, the performance drops when the distributions of training and test data differ. 
A preliminary direction is to diversify the training domains~\cite{chen2023meta,zhou2021domain} and statistically mitigate the effect of \ca{} information. Though the robust representation can be learned as the commonality across all domains,
it is implausible to exhaust the images of all possible test domains during training.

\begin{figure}
    \centering
    \includegraphics[width=0.85\linewidth]{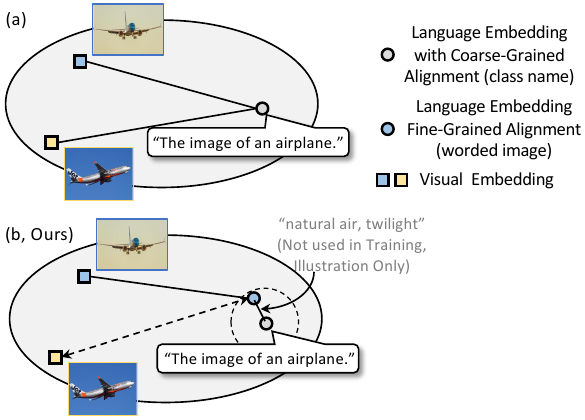}
    \caption{Given a \vlmm{} embedding space, the image and its description can be aligned. (a) However, the alignment granularity between image and its class description is at coarse-grained level where the same text can be aligned with a set of images. (b) For each image, we propose to find the \txtemb{} with fine-grained alignment, whose difference with the language embedding of class name can be used to indicate \ca{} information in the instance (\eg, twilight \& air background) and then facilitate \cs{} visual representation learning.} 
    \label{fig:concept}
\end{figure}

As an alternative, other modalities such as language have been investigated to extend vision model to unseen domains.
Recent research re-examines the image-language alignment, and use the \txtemb{s} to augment \visemb{s}~\cite{dunlap2023lads} and help train a robust classifier~\cite{cho2023promptstyler}.
In detail, the semantic difference between image description can represent the visual disparity, and then used for image synthesis from existing data.
However, as shown in Fig.~\ref{fig:concept} (a), the text can only describe concepts at coarse-grained level, such as ``The image of an airplane'', which missed visual details~\cite{ma2024mode} and can be aligned with multiple images.
As a result, the language embeddings are mainly be used to indicate different domains, which may even require precise test domain description as task priors, and cannot be used to learn robust representation for each individual image explicitly.

In this paper, we aim to learn \cs{} visual representation without any diversified domains or target priors, such that the model can overcome complex uncertainty at inference time. In particular, we study the single-source domain generalization and consider fine-grained image-language alignment where each \txtemb{} can preserve sufficient details for an image. Then, we propose the framework \approach{}, \underline{W}ording \underline{I}mage for \underline{D}omain-\underline{In}variant representation learning, and use the language model outputs to design the supervision for \cs{} representation learning.
As shown in Fig.~\ref{fig:concept}(b), we first learn the \txtemb{} which is aligned with raw visual embedding with high granularity, \ie, containing instance-specific information of the image.
In practice, as manually annotating each image with sufficient details is implausible, we represent the image as word tokens, \ie, image wording, and then extract the embedding output from the language model. 
Then, by exploring the \cs{} nature of the class name, for each image, we compare the \txtemb{s} extracted from the worded representation and the class name, and use their difference to indicate the \ca{} information. This is used to, for each image, decouple the raw visual embedding and separate the \cs{} visual representation for robust generalization. 
For simplicity, we only train a single linear layer for classification.

Through experimental validation, our \approach{} is generalizable to \vlmm{} embedding space obtained by either 1) large-scale pre-training such as CLIP~\cite{radford2021learning} or 2) contrastive training upon two uni-modal models, such as MoCo~\cite{he2020momentum} and BERT~\cite{devlin2018bert}, with the data on a single domain. 
For the data used to train CLIP encoders, it has been observed that the caption is short and can only describe partial visual content~\cite{ma2024mode,xu2023demystifying}, which will inevitably hurt the generalization to new domains~\cite{dunlap2023lads}.
For the latter one, we fix the language encoder and first train a single layer on top of the vision encoder and map the pre-trained visual embedding to the same space as the \txtemb{s} of class descriptions, where \approach{} is then applied.
Consequently, our approach provides a general solution on \vlmm{} embedding space learned from such low-quality image-language data pairs with coarse-grained alignment, and can be used to optimize the vision and language encoders pre-trained from different datasets and objectives individually.
As a summary, our contributions are as follows:
\begin{itemize}[leftmargin=*]
    \item We investigate the granularity of image-language alignment in the joint embedding space, and considers the fine-grained alignment for the problem of single-source domain generalization. 
    \item We design \approach{} to first obtain fine-grained image-language alignment and then use the language model outputs to guide \cs{} visual representation learning. This framework is general and can be built on top of encoders which are trained either jointly or separately.
    \item We experimentally demonstrate the benefit of \approach{} on three domain generalization benchmarks. 
\end{itemize}

\section{Related Work}

\paragraph{Debiased Feature Learning} has been critical to learn a robust algorithm~\cite{torralba2011unbiased} across different scenarios. In long-tail classification, the bias is caused by imbalance of data distribution and the accuracy on minor classes with limited training data is unacceptable. To mitigate the accuracy gap between minor classes and the major classes counterpart, the approaches based on data re-sampling~\cite{geirhosimagenet}, training curriculum design~\cite{zemel2013learning}, loss adjustment~\cite{kang2021exploring,menon2020long,lin2017focal} are investigated. 
Besides, the bias can also be caused by the absence of diverse domains~\cite{wang2018deep} and environments~\cite{creager2021environment}. Then, to remove such class-irrelevant features which contain domain and instance-specific information, recent approaches has applied feature disentanglement~\cite{misra2016seeing},  adversarial training~\cite{hoffman2018cycada}, and invariant regularizer~\cite{arjovsky2019invariant}. 
Besides improving the training strategy, several data pre-processing techniques~\cite{kirillov2023segment} have been used to augment the dataset and include more training data in large-scale pre-training. However, the limitation still exists as the empirical discovery cannot exhaust all training case to mitigate the biase.

\vspace{-6pt} \paragraph{Multimodal Representation} aims to combine the semantic meaning from different modalities, \eg, text and image.
With the success of self-attention~\cite{vaswani2017attention} and self-supervision such as masked modelling~\cite{devlin2018bert,he2022masked,han2022multi}, the joint encoders~\cite{sun2019videobert,tan2019lxmert,han2022few} are proposed, which take the concatenation of text and images as input and model their correlation automatically.
Recently, CLIP~\cite{radford2021learning} proposes to set encoders for each modality individually and optimize the networks through contrastive learning~\cite{chen2020simple,he2020momentum}. In this way, all extracted representation reside in the common space to be compared directly and used for down-stream application. Recently, \cite{zhai2022lit} proposes to transform the pre-trained visual embedding space into the multi-modal embedding space by aligning the language embeddings with the image embeddings extracted by the pre-trained encoder. 
However, the representation is still biased~\cite{dunlap2022using,yang2022tempclr}, and requires update via finetuning~\cite{gao2021clip} or prompting~\cite{zhou2022conditional}.

\vspace{-6pt} \paragraph{Domain Adaptation \& Generalization} aims to mitigate the performance gap caused by domain shift. To reduce the dependence on label annotation in the supervised adaptation~\cite{motiian2017unified}, the semi- and weakly-supervised setup are studied~\cite{saito2019semi,baek2020weakly} while the self-supervision is also useful~\cite{xu2019self}.
However, all of them depends on additional training data of target domains, instead, we focus on domain generalization~\cite{zhou2021domain} where only the data from source domain is used for training. Then, with the assistance of language description, the generative approaches~\cite{wang2020cross,dunlap2022using} synthesize features across domains. However, the priors of new domains are still needed. 
Recently, multi-modal prompting, \eg, CoCoOp~\cite{zhou2022conditional}, is generalizable to unseen domains by predicting classifier weights for each test image dynamically. However, it is computational expensive at test time and does not improve the quality of \visemb{}. 
Besides, the CLIP embedding space can facilitate source-free domain generalization~\cite{cho2023promptstyler} where the classifier trained on augmented language embeddings can then be used to recognize images. However, such approach heavily relies on the alignment quality of the pre-trained data.
Instead, ours is trained one single domain without any prior and aims to disentangle the \cs{} representation which is cheap at inference and robust to different domains. 
%


\section{Wording Image for Domain-Invariant Representation}\label{sec:approach}

Domain-invariant representation can be robust for test data from various domains. Due to the limitation of training data diversity in single-source domain generalization (\cref{sec:formulation}), we refer to the \txtemb{s} to describe the \cs{} visual information.
Since the short description for each image, \eg, ``an image of an airplane'', is general and can be aligned with a board set of images, the extracted \txtemb{s} can only be used to describe coarse-grained concepts for cross-domain synthesis, and cannot be used for \cs{} representation learning.

As a consequence, we choose to first learn the \txtemb{} with fine-grained alignment via image wording (\cref{sec:architecture}).
Then, we use its difference with the \txtemb{} of class name to estimate the \ca{} counterpart for visual representation disentanglement (\cref{sec:train}). Besides direct application on the joint \vlmm{} embedding space by CLIP~\cite{radford2021learning}, \approach{} is also generalized to uni-modal models such as MoCo~\cite{he2020momentum} and BERT~\cite{devlin2018bert} (\cref{sec:unimodal}).

\begin{figure*}[t]
    \centering
    \includegraphics[width=0.9\linewidth]{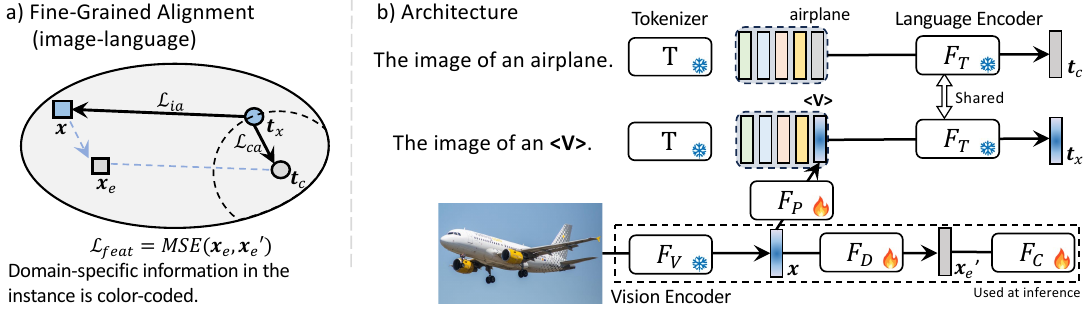}
    \caption{a) For each image, we find a \txtemb{} $\txtpseudo{}$ that is both aligned with its raw visual embedding $\mathbf{x}$ at fine-grained level and close to the \txtemb{} of class name $\txtvec{}$, under the alignment supervision at instance- ($\mathcal{L}_{ia}$) and class-level ($\mathcal{L}_{ca}$). Then, the \cs{} visual embedding $\mathbf{x}_e$ can be estimated by deducting the difference $\txtpseudo{}-\txtvec{}$ from $\mathbf{x}$ and be used as groundtruth to supervise the feature disentangler ($\mathcal{L}_{feat}$). b) For network architecture, we use the worded image token representation \texttt{<V>} for feature extraction. At training time, we fix the image \& language encoders. At test time, only the modules in dashed box are used.}
    \label{fig:approach}
\end{figure*}

\subsection{Preliminary: Single-Source Domain Generalization}\label{sec:formulation}

Domain generalization applies a model learned on the source domain to the target domains without any prior. Given the data $\srctrain{} = \{(c, \mathbf{x})\}$ for model training where $c \in \mathcal{C}$ is the class name of image $\mathbf{x}$, the model is evaluated on the test data from various domains. Formally, the test set is $\tarall{} = \big(\cup_{i=1}^{N} \tarvalid{i}\big) \cup \srcvalid{}$ where $\tarvalid{i}$ and $\srcvalid{}$ are from target domain $i$ and source domain respectively. 

Different from the multi-source domain generalization where the training data are from various domains, we consider the data from a single domain $s$. Then, for the convenience of description, we set $\srcvalid{} = \tarvalid{0}$ and then $\tarall{} = \cup_{i=0}^{N} \tarvalid{i}$. 
We note that no prior such as domain label/split is provided for test data. For fair comparison, we calculate the accuracy for each domain $\tarvalid{i}$.

As shown in Fig.~\ref{fig:approach}, for each class $c$, we only use the class name and refer to the default template in CLIP~\cite{radford2021learning}, \ie, [\texttt{The image of a/an}], to build the general description $\txtdes{}$ for all of its images. Then, we extract the \txtemb{} $\txtvec{} = \txtenc{(\txtdes{})} \in \mathcal{R}^d$ where $\txtenc{}$ is the language encoder and $d$ is the feature dimension. 
For the convenience of description, we use $\mathbf{x} \in \mathcal{R}^d$ to indicate the \visemb{} extracted by the vision encoder $\visenc{}$.

\subsection{Fine-Grained Image-Language Alignment}\label{sec:architecture}

This subsection discusses how to obtain the language embedding with fine-grained alignment for an image, \ie, $\mathbf{t}_x$ in \cref{fig:approach}(a). We first explain how to directly apply \approach{} on the pretrained joint \vlmm{} embedding space such as CLIP~\cite{radford2021learning}, while additional details regarding the generalization on uni-modal models are in \cref{sec:unimodal}.
To obtain the language embedding that preserves the image details, as annotating the full details in text could be impractical and biased, we choose to estimate the language embeddings from the images directly, which is realized by designs on both embedding alignment and architecture.

\noindent\textbf{Embedding Alignment}.
We consider two necessary conditions for fine-grained alignment: 1) be uniquely aligned with the raw visual embedding, and 2) be close with the language embeddings of the corresponding class name. 
The first condition is to, for each image, have the corresponding \txtemb{} be distinguishable and separated from the embeddings of other images, which emphasizes both \cs{} and \ca{} information in the image instance, and thus called instance-level ($\mathcal{L}_{ia}$). 
The second condition is to ensure the embedding is still discriminative and in the same distribution as the \txtemb{s} of the class names for classification, which is then called class-level ($\mathcal{L}_{ca}$).

Instance-Level Alignment $\mathcal{L}_{ia}$: the implementation follows the concept of contrastive learning~\cite{chen2020simple} and treats the \txtemb{} extracted from the image and the raw visual embedding as two views ($\txtpseudo{}, \visvec{}$), formulating a positive pair. By contrasting the positive pair from uncorrelated embeddings of other instances in the same batch (negative), the model can accurately measure the similarity of between visual embedding and the \txtemb{} with high granularity. Given the batch of images $\mathcal{B}$, the loss for each image $\visvec{} \in \mathcal{B}$ is

\begin{align}\label{eq:Con}
    \mathcal{L}_{ia}(\visvec{},\mathcal{B})  &= -\log \frac{\exp(\visvec{} \cdot \txtpseudo{} / \tau )}{\sum\nolimits_{\mathbf{x}' \in \mathcal{B}_n} \exp( \visvec{}' \cdot \txtpseudo{}  / \tau )} \\
    &- \log \frac{\exp(\visvec{} \cdot \txtpseudo{} / \tau )}{\sum\nolimits_{\mathbf{x}' \in \mathcal{B}_n} \exp( \visvec{} \cdot \mathbf{t}_{x'}  / \tau )}. \nonumber
\end{align}
where all of the embeddings are pre-processed by $l_2$-normalization and $\tau$ is a hyperparameter to rescale the affinity scores (set as 0.07 in \cite{chen2020simple}). 

Class-Level Alignment $\mathcal{L}_{ca}$ is then comparing each embedding $\txtpseudo{}$ with the language embedding of class names, and we thus apply cross-entropy loss in implementation, \ie, $\mathcal{L}_{ca}(\txtpseudo{})  = -\log \frac{\exp(\txtpseudo{} \cdot \txtvec{} / \tau )}{\sum\nolimits_{c' \in \mathcal{C}} \exp( \txtpseudo{} \cdot \mathbf{t}_{c'} / \tau )}$.

\noindent\textbf{Architecture Design}.
To ease the network training, the architecture should intuitively preserve the detailed information in the visual embeddings. Inspired by the textual inversion in image generation~\cite{gal2022image}, we think the word token projected from the visual embedding contains both \ca{} and \cs{} information. 
Thus, as shown in Fig.~\ref{fig:approach}(b), we define a two-layer MLP~\cite{lecun2015deep} as the projector $\mathbf{F}_P$ to map the \visemb{} to the token embedding. Then, we follow CLIP on how the general description $l_c$ is built, and use the same prompt [\texttt{The image of an <V>}] to get the input to $\txtenc{}$ for embedding extraction, \ie,  
\begin{equation}
    \txtpseudo{} = \mathbf{F}_T\big(\text{Concatenate}[\mathbf{T}(\texttt{The image of an}), \mathbf{F}_P(\mathbf{x})]\big)
\end{equation}
where $\mathbf{T}$ denotes the function to tokenize the input sentence and retrieve token embeddings of dimension $d$, $\mathbf{F}_P(\mathbf{x}) \in \mathcal{R}^d$ is the embedding indicated by \texttt{<V>}. 
We acknowledge that directly learning an MLP to map the visual embedding to the \txtemb{} with fine-grained alignment is an intuitive baseline. We experimentally find the image wording is more stable and better than the MLP baseline (details discussed in \cref{sec:ablation}) and thus keep it in implementation.

\subsection{Domain-Invariant Representation Learning}\label{sec:train}
Given the \txtemb{} that is aligned with the raw visual embedding with high granularity, we can then estimate the corresponding \ca{} representation and perform feature disentanglement.
In particular, as shown in Fig.~\ref{fig:approach}, we utilize the \cs{} property of class names and then consider the language embedding of class names, $\txtvec{}$, as the anchor. Then, we compare language embedding extracted from the worded image representation and class name, \ie, $\txtpseudo{} - \txtvec{}$, to estimate the \ca{} information, which is further used to denote the difference between visual embeddings~\cite{dunlap2023lads,ma2024mode} and removed from the raw visual embedding $\visvec{}$ to obtain the corresponding \cs{} visual component $\visvec{}_e$, \ie,
\begin{equation}\label{eq:global}
    \visvec{}_e = \visvec{} - k(\txtpseudo{} - \txtvec{}).
\end{equation}
where the scalar $k$ is used to align the norm of embeddings from different modalities. 
To stabilize the training, all of the embeddings are $l_2$-normalized and we set $k=1$. Given $\visvec{}_e$ as the groundtruth, we train a linear layer $F_D$ as a disentangler to predict the \cs{} feature from $\mathbf{x}$ and feed it into a linear layer $F_C$ for classification. For simplicity in implementation, we minimize the Mean Squared Error (MSE) to supervise the disentangler $F_D$, and the loss is termed as $\mathcal{L}_{feat}$. The classifier $\mathbf{F}_C$ is updated by minimizing the cross-entropy loss.

In this way, the predicted \cs{} visual embedding is directly used in evaluation. The language encoder is only used to guide the disentanglement for visual embeddings, and can be disgarded at test time to reduce computation workload. This is different from conventional prompting algorithm~\cite{jia2022visual} where the images are used to provide context and the embeddings are used for testing. 
During training, as discussed in \cref{sec:ablation}, we experimentally found that using prompt can lead to slightly better performance than directly extracting language embeddings from the class names or mapped token embeddings \texttt{<V>}. Meanwhile, our approach is robust to different prompt selections.

Finally, to avoid training collapse, we separate the training for classification from the learning of \txtemb{} with fine-grained alignment. Initially, we only train the projector $F_P$ under the alignment losses $\mathcal{L}_{ia} + \mathcal{L}_{ca}$. Then, we fix the pre-learned $\mathbf{F}_P$, and train the feature disentangler $F_D$ and the classifier $F_C$. We set batch size as 64 and the experiments can be realized on one TiTAN RTX GPU. More details can be found in Appendix.

\subsection{Generalization on Uni-Modal Models}\label{sec:unimodal}

Besides the joint \vlmm{} embedding space learned via large-scale pre-training, it can also be obtained by contrasting the class names with the training images $\srctrain{}$ in a single domain.

Specifically, we extract the language embeddings of class names such as ``an image of an airplane'', as class centers. For each image, its visual embedding should be close to the language embedding of corresponding class name and distant from others. In practice, as the inputs to the language encoders are limited, we choose to fix the language encoder to avoid overfitting, and project the visual embedding space to be aligned with the pre-trained language embedding space. To reduce the training cost further in the image encoder update, we train a linear layer on top of the fixed image encoder. In this way, we essentially use the space of language embeddings as a joint \vlmm{} embedding space and then apply \approach{} to use \txtemb{} as the \cs{} anchor to instruct the disentanglement of the \visemb{}.

We acknowledge the training data in estimating this joint embedding space is limited and the pairing between image and its class description is at coarse-grained level, compromising the effect of learned space. However, as pointed in MoDE~\cite{ma2024mode}, for the data used in CLIP pre-training, similar issue of image-language alignment still exists, \ie, the caption is short and each caption can only describe partial visual content in an image. 
As a summary, our approach provides a general framework for visual embedding disentanglement on a joint \vlmm{} embedding.

\section{Experiment}\label{sec:da}

We explain the experimental evaluation (\cref{sec:eval-setup}) on the \vlmm{} embedding space learned via 1) large-scale pretraining and 2) aligning uni-modal models for data in a single domain separately (\cref{sec:main-experiment}). Then, we ablate the approach design (\cref{sec:ablation}). Moreover, we extend \approach{} in long-tail learning and provide details in the \cref{sec:tail}.

\subsection{Evaluation Setup}\label{sec:eval-setup}

\vspace{3pt} \bitem{Dataset.} 
1) \textit{CUB-Painting} consists of 200 finegrained classes of bird species from two domains constructed upon CUB-200-2011~\cite{wah2011caltech} and CUB-200-Paintings~\cite{wang2020progressive}. The former one (source) contains the pictures taken from natural world, while the latter one (target) is in the painting style. 2) \textit{DomainNetMini}~\cite{tan2020class} is a subset of DomainNet~\cite{peng2019moment} and contains 40 classes from four different domains, \ie, sketch, real, clipart and painting. We follow the commonly used evaluation setup~\cite{dunlap2023lads} where the model is only trained on ``sketch'' domain (source) and the rest three domains (target) are unseen during training. 3) \textit{Office-Home}~\cite{venkateswara2017deep} contains 65 classes from four different domains, \eg, real, art, clipart, product. We train the model on the real domain (source) and the rest three domains (target) are unseen during training.

\vspace{3pt} \bitem{Evaluation Metric \& Implementation.} We report the accuracy on the source (\id{}) and target (\ood{}) domains respectively, and use the average (\all{}) to measure the performance on all domains. 

\subsection{Experimental Study}\label{sec:main-experiment}

\begin{table*}[]
\centering
\caption{Performance comparison of domain generalization.}
\resizebox{0.8\linewidth}{!}
{\renewcommand{\arraystretch}{1.}
    \begin{tabular}{c|ccccccccc}
    \hlineB{3}
    \multirow{2}{*}{Approach} & \multicolumn{3}{c}{CUB-Painting}  & \multicolumn{3}{c}{DomainNetMini}  & \multicolumn{3}{c}{Office-Home}\\
                             & \id{}    & \ood{}   & \all{} & \id{}    & \ood{}   & \all{}  & \id{}    & \ood{}   & \all{} \\ \hline
    Zero-Shot~\cite{radford2021learning}                & 60.34 & 52.84 & 56.59    & 93.49   & 95.94   & 94.72  & 92.20 & 82.54 & 84.37  \\
    Zero-Shot*~\cite{radford2021learning}               & 61.93 & 54.38 & 58.16    & 93.24   & \textbf{96.01}  & 94.62 & 95.00 & 85.46 & 90.23    \\
    WiSE-LD~\cite{wortsman2022robust}                  & 81.74 & 64.80 & 73.27    & 95.19   & 93.68   & 94.44   & -     & -     & -   \\
    VQGAN+CLIP~\cite{crowson2022vqgan}               & -     & -     & -        & 95.54   & 93.83   & 95.27  & -     & -     & - \\
    LADS~\cite{dunlap2022using}                     & 86.14 & 66.18 & 76.16    & 95.33   & 95.21   & 95.27   & -     & -     & - \\ 
    Linear Clf.              & 85.91 & 64.33 & 75.12    & 95.03   & 93.75   & 94.39    & 95.65 & 84.30 & 89.97 \\
    MLP Clf.                & 86.06 & 62.32 & 74.19 & 96.33 & 92.47 & 94.40  & 95.70 & 82.87 & 89.04 \\ \hline
    \approach{} (Ours)                     & \textbf{87.47} & \textbf{68.76} & \textbf{78.12}    & \textbf{97.08}   & \textbf{96.01}   & \textbf{96.75}  & \textbf{96.34} &  \textbf{87.35} & \textbf{91.84} \\
    \hlineB{3}
    \multicolumn{10}{l}{{*: include domain in extracting \txtemb{} for each class.}} \\
    \end{tabular}
}\label{tab:domain}
\end{table*}

\vspace{3pt} \bitem{Generalization on Large-Scale Pre-trained Models.} For fair comparison, we use pre-trained OpenAI CLIP~\cite{radford2021learning} ViT-L/14 for study and in effect use the CLIP language embeddings to help with the disentanglement of CLIP visual embedding.
As summarized in Table~\ref{tab:domain}, our approach achieves the best scores on all three benchmark dataset consistently. 

Admittedly, CLIP has shown remarkable ability to generalize across different tasks. However, learning a linear classifier (Linear Clf.) on the CLIP visual embeddings is still necessary to improve the performance significantly as using \txtemb{s} of classes as centers (Zero-Shot) cannot clearly identify the semantic details for fine-grained objects on CUB-Painting. Nevertheless, Linear Clf. may show bias to the source domain and the bias will be more significant when more parameters are learned. On CUB-Painting, there is a clear performance gap between Src. and Tar. On DomainNetMini, Linear Clf. may even underperforms the zero-shot counterparts on Tar. In addition, by employing a stronger classifier (MLP Clf.), which includes two layers but is only trained by minimizing cross-entropy loss, the accuracy on Tar. drops further, resulting larger performance gap.
In contrast, our approach learns \cs{} visual representation and can outperform all other baselines including LADS consistently on all cases but does not need domain prior.

\bitem{Generalization on Aligned Uni-Modal Models.}
We evaluate the generalization of \approach{} by aligning four image encoders and two language encoders. For each image encoder, we use the pre-defined data pre-processing function. As mentioned in \cref{sec:unimodal}, we only train a linear layer to map the \visemb{s} and both of the two encoders are fixed. Thus, as a fair comparison, for the baseline Linear Clf., this mapping layer is also added and updated jointly with the classifier. More implementation details can be found in the Supp. 

As summarized in \cref{tab:uni_fixed}, we use CUB-Painting to experimentally validate the effect of \approach{} for fine-grained recognition in single-source domain generalization. 
We select the public available image encoders trained in Full-Supervision~\cite{lecun2015deep}, DINO~\cite{caron2021emerging}, MoCo~\cite{he2020momentum}, and MAE~\cite{he2022masked}, with different architectures if applicable. All of the models are pre-trained on ImageNet~\cite{deng2009imagenet} and the pre-training approaches are arranged according to the linear probing accuracy on ImageNet test set in an decreasing order, which is used to indicate how discriminative the visual embeddings are. Then, we consider the language encoder trained in CLIP~\cite{radford2021learning} (CLIP-Text) or BERT~\cite{devlin2018bert} where the output corresponding to class token \texttt{<CLS>} is the \txtemb{} to be aligned. 

\begin{table*}[t]
    \centering
    \caption{Performance Comparison on CUB-Painting via Uni-Modal Models Alignment}
    \resizebox{0.8\linewidth}{!}{
    \begin{tabular}{c|cc|cc|cc|cc}
    \hlineB{3}
                & \multicolumn{2}{c|}{Fully-Supervised~\cite{lecun2015deep}} & \multicolumn{2}{c|}{~~~~~~DINO~\cite{caron2021emerging}~~~~~~} & \multicolumn{2}{c|}{~~~~~~MoCo~\cite{he2020momentum}~~~~~~} & \multicolumn{2}{c}{~~~~~~MAE~\cite{he2022masked}~~~~~~} \\  \hline \hline 
     & \multicolumn{2}{c|}{(ViT-B$_{16}$)}  & \multicolumn{2}{c|}{(ViT-B$_{16}$)} & \multicolumn{2}{c|}{(ViT-B$_{16}$)} & \multicolumn{2}{c}{(ViT-B$_{16}$)} \\
        & ~~~~Src.              & Tar.             & ~~~Src.        & Tar.       & ~~~Src.       & Tar.       & ~~~Src.        & Tar.       \\ \hline
    Linear Clf. & ~~~~66.29             & 37.74          & ~~~80.45       & 52.67     & ~~~53.18       & 25.66      & ~~~32.03      & 31.99         \\
    CLIP-Text   & ~~~~70.31             & 35.08       & ~~~81.34       & 53.36          & ~~~75.49       & 39.51      & ~~~54.42      & 48.08       \\
    BERT        & ~~~~75.6              & 41.25          & ~~~81.71       & 54.05         & ~~~72.94       & 35.77      & ~~~51.07      & 42.87     \\ 
    \hline \hline 
      & \multicolumn{2}{c|}{(RN50)}  & \multicolumn{2}{c|}{(ViT-B$_{8}$)}  & \multicolumn{2}{c|}{(RN50)} & \multicolumn{2}{c}{\multirow{2}{*}{-}} \\
     & ~~~Src.              & Tar.             & ~~~Src.        & Tar.     & ~~~Src.        & Tar.       &       &         \\ \hline
    Linear Clf. & ~~~50.12 & 22.32 & ~~~80.98 & 49.66 &  ~~~49.52 & 14.17 & &  \\
    CLIP-Text   & ~~~51.85 & 21.6 & ~~~82.62 & 52.41 & ~~~53.4  & 19.4  &  &   \\
    BERT        & ~~~68.67 & 26.49 & ~~~82.62 & 52.84 & ~~~49.53 & 14.67 &  &   \\
    \hlineB{3}
    \end{tabular}}
    \label{tab:uni_fixed}
\end{table*}

First of all, we discuss the combination of the uni-modal encoders and we found the benefit of \approach{} is better explored when the pre-training objectives of the encoders are complementary to each other. In detail, BERT provides more gain for the vision encoder pretrained by DINO and Full Supervision while CLIP-Text benefits more for the other two models. 
\begin{itemize}[leftmargin=*]
    \item The embeddings learned in Full-Supervision and DINO is more discriminative and may focus more on global semantic meaning. Then, as BERT is trained with multiple tasks including word prediction that acquires local semantic modeling, it can better locate the \ca{} component exhibited in the \visemb{} to facilitate the feature disentanglement. 
    \item MoCo and MAE can preserve more local details in pre-training, \ie, MoCo is essentially performing pair-wise comparison across instances and MAE is trained to reconstruct the masked area. Then, as CLIP-Text is learned under the contrastive objective to cluster features with high semantic similarity, the \txtemb{s} of classes can serve as desired \cs{} anchor so that the \ca{} details in each image can be better represented.  
\end{itemize}

As such, when compared with Linear Clf. baseline, though applying \approach{} can improve the accuracy on both source and target domains in most cases, exception still exists, \ie, aligning CLIP-Text and the vision encoder pre-trained by Full-Supervision. As CLIP-Text is trained with contrastive loss, it may not capture the tiny details in the visual data to facilitate the \cs{} disentanglement. Furthermore, since the training data of ImageNet is also captured in natural world, the vision encoder may overfit to source domain such that the \visemb{s} in target domain have already been distant from those of natural images. As such, the classifier learned from the \cs{} component in the source domain cannot be generalized to the target domain.

Regarding the architectures, for each vision encoder, we consider both Transformer and ResNet if the checkpoints are available. As ViT-B is of more learnable parameters than RN50 (86M Vs. 23M), ViT-B outperforms RN50 consistently. Meanwhile, we note the benefit of our approach is better explored on larger models. For example, for MoCo~\cite{he2020momentum}, the performance gain with ViT-B on Src. and Tar. are 21.03 and 11.98 respectively, but are 1.95 and 2.86 with RN50. For DINO, from the Linear Clf., the model with larger patch size can facilitate generalization on target domains but has slightly weaker score on source domain. However, \approach{} can consistently improve the performance and minimize the gap between ViT-B$_{16}$ and ViT-B$_{8}$,

Finally, comparing with results in \cref{tab:domain}, though the vision encoder in CLIP still achieve the best performance, we note the pre-train set of CLIP is much larger than ImageNet. And we believe the performance can still be further improved if a stronger pre-trained vision encoder is provided. 

In sum, as learning a large-scale \vlmm{} model is expensive in both data collection and network training, and such a giant model does not always outperform a smaller one specifically trained in one modality, \eg, the vision encoder in CLIP underperforms the backbone by MAE~\cite{he2022masked} for object detection, we hope \approach{} provides more insights for aligning uni-modal models for different downstream applications.

\subsection{Ablation Study}\label{sec:ablation}

This subsection first evaluates the \approach{} design including prompt, alignment, and necessity of image wording. Then, the properties of the disentangled embeddings and \txtemb{s} extracted from the image are analyzed.

\vspace{3pt} \bitem{Robustness to prompts.} \approach{} currently uses [The image of a/an \texttt{<V>}] as the prompt to extract language embeddings, but can be applied to different prompts and even trained among a mixture of prompts to improve the accuracy further. We additionally consider other two prompts, [The photo of an \texttt{<V>}] and [\texttt{<V>} in the scene], and use CUB-Painting for validation.
First of all, comparing Table~\ref{tab:domain} \&~\ref{tab:ablation}(a), when randomly selecting a prompt for each image (\textit{Random}) during training, the resulted score is similar. Next, we aggregate the embeddings extracted from multiple prompts (\textit{Aggregated}). Formally, we gather outputs $\{\txtvec{(i)}\}_{i=1}^N$ and $\{\txtpseudo{(i)}\}_{i=1}^N$ where $i$ indexes the prompt and $N=3$. As a result, we treat $\bar{\txtvec{}} = \frac{1}{N} \sum_i \txtvec{(i)}$ and $\bar{\txtpseudo{}} = \frac{1}{N} \sum_i \txtpseudo{(i)}$ as language embeddings at coarse- and fine-grained alignment for disentanglement. As more prompts are used to stabilize the training, the score is improved. Meanwhile, to correctly estimate the \ca{} feature, we note the prompt used to extract the pair of embeddings ($\txtvec{},\txtpseudo{}$) for each image should be the same. Otherwise, the difference between prompts may introduce noises (\eg, $\txtpseudo{(2)}-\txtvec{(1)}$) and result in slight drop (\textit{Misaligned}). Finally, we observe that directly existing language embedding without prompt (\textit{None})  leads to slight performance drop. One potential reason is that the meaning of a single class name could be vague and using prompt can provide background as context.

\bitem{Instance-Level Alignment} is necessary to learn the \txtemb{} with fine-grained alignment, and minimizing the contrastive loss (CT) is important to distinguish embeddings from different instances. As an additional evidence, in Table~\ref{tab:ablation}(b), we apply supervised contrastive loss (SupCT) and match the \txtemb{} extracted from worded image tokens with the raw visual embeddings from different images. As the resulted alignment is still at coarse-grained level, the generalization on target domain drops. Similarly, when no instance-level alignment (None), the \ca{} details in each instance cannot be properly estimated either.

\begin{table*}[]
    \caption{Ablation study for (a) Robustness to prompts and (b) Instance-level alignment regularization. (c) Classification accuracy of language embeddings with fine-grained alignment.}
    \parbox{.33\textwidth}
    {
        \centering
        \resizebox{.3\textwidth}{!}
        {%
            \renewcommand{\arraystretch}{1.0}
            \begin{tabular}{ccc}
                \hlineB{3}
                   (a)   & \id{}    & \ood{}  \\ \hline
                Random     & 87.49 & 68.84 \\
                Aggregated & 88.51 & 70.01 \\
                Misaligned & 87.13 & 68.20 \\
                None & 86.95 & 67.83 \\ 
                \hlineB{3}
            \end{tabular}
        }
    }
    \hfill
    \parbox{.3\textwidth}
    {
        \centering
        \resizebox{.26\textwidth}{!}
        {%
            \renewcommand{\arraystretch}{1.2}
            \begin{tabular}{ccc}
                \hlineB{3}
                (b)    & \id{}    & \ood{}  \\ \hline
                None  & 86.42 & 67.12   \\
                SupCT & 87.11 & 67.54   \\
                CT    & \textbf{87.47} & \textbf{68.76} \\
                \hlineB{3}
            \end{tabular}
        }
    }
    \hfill
    \parbox{.35\textwidth}
    {
        \centering
        \resizebox{.28\textwidth}{!}
        {%
            \renewcommand{\arraystretch}{1.0}
            \begin{tabular}{ccc}
            \hlineB{3}
            (c) & ~~Src. & Tar.   \\ \hline
            & \multicolumn{2}{c}{CUB-Painting} \\
            & ~~~~84.43~~~ & ~~~59.76~~ \\ \hline
            & \multicolumn{2}{c}{DomainNetMini} \\
            & ~~~~97.00~~~ & ~~~93.86~~  \\ 
            \hlineB{3}
            \end{tabular}
        }
    }\label{tab:ablation}
\end{table*}

\bitem{Estimation of \txtemb{} with fine-grained alignment} is realized via image wording in \approach{}. As a naive baseline, we can learn an MLP layer to predict the embedding under the alignment regularization. However, the performance is sub-optimal (CUB-Painting, \id{} 63.77 and \ood{} 48.98). After all, the parameters learned in the language encoder can also determine the distribution of language embedding. In other words, even in the joint embedding space, there is still gap between the embedding distribution of different modalities. Then, purely training a simple MLP on limited data to predict the embedding with high granularity is difficult.
Similarly, though the \txtemb{} of class names is naturally \cs{}, learning to directly map the raw \visemb{} to this anchor is more difficult and easier to overfit when training on a single domain. In contrast, \approach{} explicitly learns to represent the \ca{} information and is more robust across domains.

\bitem{Necessity of learning \cs{} visual representation.} As the \txtemb{} extracted from worded image is trained under alignment constraint, it can also be used for evaluation. However, as the language encoder is fixed, the performance highly depends on the separation between pre-learned language embeddings.
By comparing Table~\ref{tab:domain} \& \ref{tab:ablation}(c), regardless the slight gain against zero-shot baseline, The Tar. accuracy still under-performs the Linear Clf. baseline clearly and the computation workload at test time is much larger.
Thus, the visual encoder is essential, and \approach{} does not forget the discriminative details but only remove the \ca{} components.
Similarly, though both the language embedding of class names $\txtvec{}$ and the disentangled component $\visvec{}_e$ are \cs{}, as the embedding distribution of different modalities still varies, directly predicting $\txtvec{}$ from the image instance is more lean to overfit to the domain bias.

\bitem{Validation of \cs{} property.} Under the guidance of language embedding, the \cs{} representation $\visvec{}_e'$ is predicted from the raw \visemb{} $\visvec{}$. 
Meanwhile, the difference between \txtemb{s} aligned with the same image but with different granularities, \ie, $\mathbf{t}_x - \mathbf{t}_c$, is used by \approach{} to represent the \ca{} information. Thus, we compare these three embedding types and evaluate the accuracy for domain generalization and domain prediction. 

\begin{table}[t]
    \caption{Performance comparison of disentangled features in (a) domain generalization and (b) domain prediction.}
    \parbox{.5\textwidth}
    {
        \centering
        \resizebox{.5\textwidth}{!}
        {%
            \begin{tabular}{c|cc|cc}
                \hlineB{3}
                \multirow{2}{*}{(a)} & \multicolumn{2}{c|}{CUB-Painting} & \multicolumn{2}{c}{DomainNetMini} \\
                                  & Src.            & Tar.           & Src.            & Tar.            \\ \hline
                Full               & 85.91           & 64.33          & 95.03           & 93.75           \\
                Domain-Specific   & 51.78           & 36.82          & 37.16           & 30.97           \\
                Domain-Invariant & 87.47           & 68.76          & 97.08           & 96.01           \\
                \hlineB{3}
            \end{tabular}
        }\label{tab:object_cls}
        
    }
    \hfill
    \parbox{.5\textwidth}
    {
        \centering
        \resizebox{.5\textwidth}{!}
        {%
            \renewcommand{\arraystretch}{1.0}
            \begin{tabular}{ccc}
                \hlineB{3}
                 (b)                & CUB-Painting & DomainNetMini \\ \hline
                Full                  & 90.85        & 76.22         \\
                Domain-Specific      & 97.31        & 88.91         \\
                Domain-Invariant    & 75.30        & 48.14         \\
                \hlineB{3}
            \end{tabular}
        }\label{tab:alignment}
    }\label{tab:disentangle}
\end{table}

Different from domain generalization where the model rely on \cs{} information to maintain high accuracy in recognizing object category on various domains at test time, domain prediction trains a model on data from all domains and then utilize the \ca{} information to predict the domain at test time. As shown in Table.~\ref{tab:disentangle}, we use CUB-Painting and DomainNetMini for evaluation and more implementation details are provided in the Supp.

As \ca{} information includes background, it thus achieves higher accuracy in domain prediction. Then, comparing with the raw visual embedding (Full), as the \ca{} counterpart has been deducted, the \cs{} feature is more effective in recognizing object categories for data from various domains but less effective in predicting the domain. 

\bitem{Limitations and societal impacts}. As introduced in Sec.~\ref{sec:approach}, we estimate the \ca{} information by comparing embeddings extracted from the language encoder. Therefore, the estimation relies on the quality of the language model. Also, \approach{} takes the output corresponding to \texttt{<CLS>} token as the language embeddings, which does not intuitively apply to autoregressive language models where no \texttt{<CLS>} token is learned. We will leave it for future work. To the best of our knowledge, as our work is purely an algorithm for learning \cs{} features, we haven not found any negative societal impact.

\section{Conclusion}

In this paper, we investigate the problem of single-source domain generalization without any test prior, which is important for real-world application. We re-exiamine the 
We propose the framework \approach{}: by aligning the distribution of visual embeddings with the distribution of language embeddings with high granularity, the difference of \txtemb{s} between the worded image representation and the class names is used to estimate the \ca{} information for each image. Ablation study demonstrates that the \txtemb{} obtained by image wording is aligned with higher granularity and is most helpful in feature disentanglement.
In this way, we distill the knowledge in language models to facilitate the \cs{} representation learning and our approach can be generalized even though the encoders are learned on irrelevant datasets separately. 
We plan to adapt WIDIn for models in advanced tasks such as object detector and generative models in the future.

{
    \small
    \bibliographystyle{cvpr2024/ieeenat_fullname}
    \bibliography{main}
}

\newpage

\newpage

\appendix


\section{Extension to Long-Tail Image Classification}\label{sec:tail}

In addition to the \ca{} information, the background details such as sea and desert in the same domain may also confuse the classification and be sensitive to distribution bias. As such, removing those information can make the model easily obtain well-separated features. In this way, \approach{} can be helpful for long-tail learning and balance the classification accuracy on all classes.

As shown in Table~\ref{tab:longtail}, we evaluate our approach on \textbf{ImageNet-LT}~\cite{liu2019large} and \textbf{iNaturalist}~\cite{inast18}, where the number of training samples per class ranges in [5,1280] and [2,1000] respectively. 
\textbf{ImageNet-LT}~\cite{liu2019large} is a long-tailed subset of the ImageNet~\cite{deng2009imagenet}. There are in total 115.8K images from 1000 categories.
\textbf{iNaturalist}~\cite{inast18} contains around 437.5K images from 8142 categories and the number of training instances per class ranges from 2 to 1000. 
According to the data size of each class, we split the classes into three groups, Many ($\geq 100$ samples), Medium (Med, 20$\sim$100 samples) and Few ($\leq 20$ samples). Then, for evaluation, we report the performance on all classes as well as the accuracy over the groups. 
For both ImageNet-LT and iNaturalist datasets, the distribution of test sets is uniform where the numbers of test data per class are 50 and 3 respectively.
Meanwhile, for our approach, we report the performance with both RN50 and ViT-B$_{16}$ as the backbone and use pre-trained CLIP model as initialization for fair comparison.

\begin{table*}[!htb]
\centering
\caption{Performance comparison of long-tail classification.}
\resizebox{0.85\linewidth}{!}
{\renewcommand{\arraystretch}{1.03}
    \begin{tabular}{cc|cccc|cccc}
    \hlineB{3}
    \multirow{2}{*}{Approach} & \multirow{2}{*}{Backbone} & \multicolumn{4}{c|}{ImageNet-LT}  & \multicolumn{4}{c}{iNaturalist} \\
                              &                           & Few   & Med & Many  & Overall & Few   & Med & Many  & Overall     \\ \hline
    LWS~\cite{kangdecoupling} & RN50$^*$                  & 29.30 & 45.20  & 57.10 & 47.70   & 65.50 & 66.30  & 65.00 & 65.90       \\
    LogitAdj~\cite{menon2020long}                  & RN50$^*$                  & 49.94 & 52.32  & 50.06 & 51.13  & 66.27 & 66.34  & 66.79 & 68.44       \\
    TSC~\cite{li2022targeted}                       & RN50$^*$                  & 30.40 & 49.70  & \textbf{63.50} & 52.40  & 67.80 & 70.60  & 72.60  & 69.70       \\
    NCM~\cite{kangdecoupling} & RN50$^{\wedge}$          & 31.10 & 46.60  & 58.90 & 49.20   & \multicolumn{3}{c}{-} & 65.30       \\
    cRT~\cite{kangdecoupling} & RN50$^{\wedge}$          & 27.80 & 47.20  & 63.30 & 50.80   & \multicolumn{3}{c}{-} & 69.90       \\
    $\tau$-normalized~\cite{kangdecoupling} & RN50$^{\wedge}$ & 33.80 & 48.40  & 60.90 & 51.20  & \multicolumn{3}{c}{-}  & 71.20       \\
    LWS~\cite{kangdecoupling} & RN50$^{\wedge}$          & 31.80 & 48.60  & 62.20 & 51.50   & \multicolumn{3}{c}{-}  & 71.00       \\
    Zero-Shot & RN50$^{\dagger}$& 52.92 & 57.78  & 58.13 & 56.61  & 2.91  & 3.43   & 3.48  & 3.41        \\
    Baseline                  & RN50$^{\dagger}$          & 53.05 & 60.24  & 59.17 & 58.85   & 69.90 & 71.81  & 73.51 & 72.60       \\
    \approach{} (Ours)        & RN50$^{\dagger}$          & \textbf{55.12} & \textbf{61.26}  & 60.56 & \textbf{60.55}  & \textbf{71.68} & \textbf{73.03}  & \textbf{73.56} & \textbf{73.20}      \\ \hline
    Zero-Shot                 & ViT-B16$^{\dagger}$       & 63.63 & 66.43  & 66.86 & 66.71   & 3.66  & 4.35   & 4.16  & 4.16        \\
    Baseline                  & ViT-B16$^{\dagger}$       & 67.84 & 67.72  & 68.23 & 67.95   & 72.11 & 73.77  & 73.97 & 73.72       \\
    \approach{} (Ours)        & ViT-B16$^{\dagger}$       & \textbf{68.91} & \textbf{71.69}  & \textbf{72.07} & \textbf{71.93}   & \textbf{75.01} & \textbf{75.00}  & \textbf{75.75} & \textbf{75.45}        \\
    \hlineB{3}
    \multicolumn{10}{l}{\small $^{\wedge}$: Initialized with CLIP~\cite{radford2021learning} and jointly trained. Results reported by \cite{tian2022vl}.} \\
    \multicolumn{10}{l}{\small $^{*}$: Trained from scratch. $^{\dagger}$: \emph{Fixed} and initialized with CLIP~\cite{radford2021learning}.}
    \end{tabular}
}\label{tab:longtail}
\end{table*}

\begin{table}[]
    \centering
    \caption{Comparison on long-tail learning.}
    {%
        \renewcommand{\arraystretch}{1.13}
        \begin{tabular}{ccccc}
            \hlineB{3}
            Approach  & Few          & Med   & Many  & STD        \\
            \hline
            Zero-Shot & 63.63        & 66.43 & 66.86 & 1.75 \\
            VL-LTR~\cite{tian2022vl}    & 59.30 ($\downarrow$ 4.33) & \textbf{74.6}  & \textbf{84.5}  & 12.70 \\
            Ours      & \textbf{68.91 ($\uparrow$ 5.28)} & 71.69 & 72.07 & \textbf{1.72} \\
            \hlineB{3}
        \end{tabular}
    }
    \label{tab:tail-compare}
\end{table}

Though CLIP shows strong zero-shot accuracy on ImageNet-LT, it does not perform well on fine-grained classes in iNaturalist. We note that most approaches compared below requires the update of vision encoder parameters to achieve high score, which is computational expensive. Instead, we fix the encoders but add class-specific margins~\cite{menon2020long,ma2023digeo} in calculating the cross-entropy loss (Baseline) and \approach{} can still achieve consistent performance gain. We believe that our approach can further improve the performance when the vision encoder is also jointly trained with the projector.

During model training, as we fix the CLIP vision encoder during the entire training process, we follow LogitAdj~\cite{menon2020long} to add class-based margins to the loss during training ans also add one linear layer with the same input-output dimension to project \& adjust features. The class margins are calculated based on the data distribution and first theoretically studied in LogitAdj~\cite{menon2020long}.

Meanwhile, as listed in Table~\ref{tab:tail-compare}, comparing with VL-LTR~\cite{tian2022vl} which update the backbone and use extra dataset for training, we fix the vision encoder after being initilized with the CLIP-pretrained model and do not need access to any extra instance-specific description, instead, we just build description for each class.
Additionally, since the visual encoder is updated, VL-LTR~\cite{tian2022vl} achieves significant performance on the Many group at the expense of sacrificing accuracy on Few group, comparing with zero-shot baseline. In contrast, our approach is capable of achieving consistent performance gain across all class groups and also reducing the accuracy variance  across these three groups. Meanwhile, we believe \approach{} is orthogonal with VL-LTR and other similar approaches such as Simple~\cite{ma2021simple}, and can be combined with it to further improve the performance.

\section{Clarification of \approach{}}

\bitem{Motivation}. Contrastive Language-Image Pre-training aims to align each image with its paired caption. In the joint \vlmm{} embedding space, by aligning a vast amount of image-caption pairs, as a necessary condition, the distribution of visual embeddings have been aligned with the distribution of language embeddings. Then, for two image-caption pairs, we assume the difference between the language embeddings can be used to estimate the difference between the image embeddings roughly.

\bitem{Alignment Granularity}.
We use the language embeddings at different alignment granularity level for domain-invariant visual representation learning. First of all, the alignment between an image and its corresponding class description is at coarse-grained level. The same text can be aligned with a broad set of images of the same class. Instead, the language description with fine-grained alignment should uniquely and specifically describe the full details in the image, but is difficult to obtain. As such, \approach{} is proposed to use image to directly estimate the aligned language embedding with high granularity. As the language embedding with high granularity is not directly predicted from the words, we thus term it as pseudo language embedding when space is limited.

\bitem{Alignment Granularity for Domain-Invariant Representation} As the language embedding are extracted from the class names and the definition of class names does not depend on any domain information, the language embeddings are domain-invariant naturally. 
Thus, for each image, we propose to use the difference between language embeddings at different alignment granularity level, to indicate the domain-specific information in the image, which is to be deducted from the visual embedding.
Meanwhile, we note the word token is predicted from the visual embedding to be disentangled instead of the raw image, because the visual encoder has already filtered out domain-specific information in the raw image and our focus is to disentangle the trained embeddings.

\bitem{Classification Granularity} is only mentioned for downstream evaluation on CUB-Painting and the model is supposed to recognize the bird of different sub-species. 
Though the CLIP model trained on large dataset such as LAION has shown strong zero-shot accuracy on different downstream tasks. As demonstrated in Table 1, the visual encoder is still not robust enough for images from different domains, in particular for fine-grained image classification.
Meanwhile, we note the alignment between the bird image with bird subspecies class name is still at coarse-grained level.

\bitem{Importance of \approach{}}. As emphasized in the Sec. 1, \approach{} provides a general solution on vision-language embedding space learned from low-quality data where the low quality means the alignment of image-caption pair in the CLIP training data has low granularity. As discussed in Sec. 4.2, we consider two cases of joint \vlmm{} embedding space, 1) CLIP embedding space which is obtained through multi-modal contrastive learning on LAION 400M, and 2) finetuning from two uni-modal models which are trained on irrelevant large-scale uni-modal datasets through multi-modal contrastive learning on the training set in source domain. As pointed in MoDE~\cite{ma2024mode}, the caption in each pair is short and only capture partial visual content, which is similar to the aligning of two uni-modal models where we use class description to finetune via contrastive learning. 

\bitem{Baseline.} \approach{} use a linear layer $F_P$ to predict the \cs{} visual representation from the original image embedding while another linear layer $F_C$ is used as the classifier upon the \cs{} visual embedding. Though the \cs{} visual embedding is still supposed to be in the same space as the raw visual embeddings, more parameters are indeed introduced. As such, as shown in Table 1, we consider both Linear Clf. and MLP Clf. as the baselines. 
The Linear Clf. is used to compare the classification accuracy obtained via raw visual embedding and the \cs{} visual embedding and our \approach{} can achieve better generalization across domains. 
Then, MLP Clf. has the same number of parameters (as well as architecture) as \approach{} at test time and the only difference is the regularization by $\mathcal{L}_{feat}$. Then, we can see that MLP Clf. even enlarges the performance gap between Src. and Tar. In contrast, \approach{} can improve the performance on both Src. and Tar. domains and mitigate the performance gap.


\section{Visualization \& Discussion}

\subsection{Comparison between vision and language embeddings}

As language embedding with high granularity are learned under both the class-level and instance-level alignment, they can thus be also used for evaluation by comparing with the \txtemb{s} extracted from the class descriptions. As shown in Table 3(c), even though the performance is improved, we found that it is still significantly under-perform the linear evaluation baseline trained on the \visemb{s}. 

As such, we think the distribution of \txtemb{s} and the distribution of \visemb{s} may not be perfectly aligned and the \visemb{s} are capable to capture tiny details which are important for fine-grained classification but not captured by the \txtemb{s}. Towards this end, from the perspective of comparison, we believe it is unfair to only compare the zero-shot performance with any approach trained on \visemb{s}. Instead, the linear evaluation should also be considered and accompanied with the zero-shot baseline for reference.

\subsection{Visualization}

As shown in Fig.~\ref{fig:visualiation} below, we provide two visualization examples of the embeddings. Within each example, we plot the \visemb{s}, \mapemb{s}, and the \cs{} \visemb{s} of two instances (the cycle shape denotes the instance from source domain and the square shape denotes the instance from the target domain), as well as the \txtemb{s} of class description. In both examples, comparing with the \visemb{s} of the two instances, the \cs{} \visemb{s} are closer to each other. In this way, the classifier learned on the instances of source domain can also perform well on the instances from target domains.
\begin{figure}
    \centering
    \includegraphics[width=\linewidth]{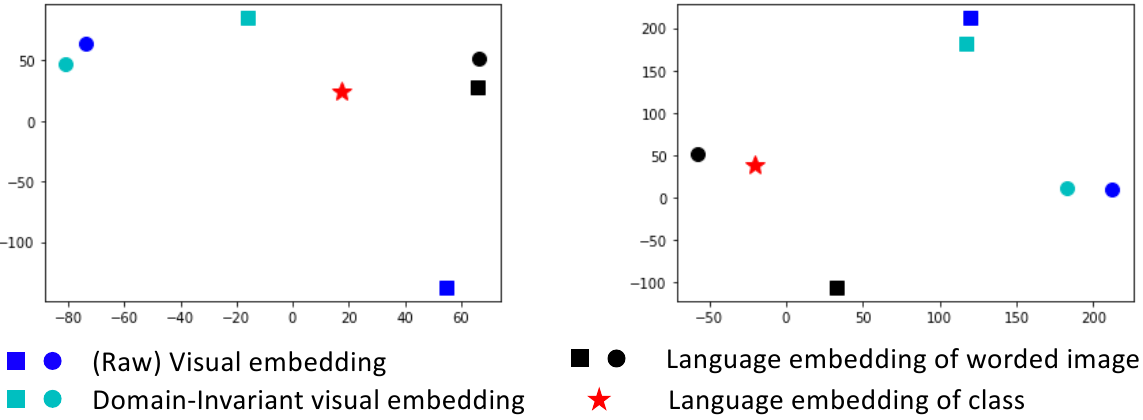}
    \caption{Visualization of embedding of two examples, the embedding types are color coded.}
    \label{fig:visualiation}
\end{figure}

\subsection{Generalization to Multi Instances}
Our approach can be naturally generalized to images containing multiple classes, as it leverages the difference between language embedding from worded image and language embedding of caption to model the class-invariant information. For example, given the caption ``The image of a cat and a dog'' and its paired image, as the original visual embedding has encoded the information from both the cat and the dog, the corresponding language embedding with high-granularity can still be compared with the language embedding of paired caption to estimate the information of background and domain.

At the same time, to model the \cs{} information for each object in the image, we can extend the prompt for single concept to the prompt [The image of a \texttt{<V$_1$>} and a \texttt{<V$_2$>}].

Then, we can follow~\cite{carion2020end} to design the projector where the visual embedding can be mapped to different word tokens. As the classes mentioned in the caption is commutative, \eg, ``The image of a cat and a dog'' is equivalent to ``The image of a dog and a cat'', and the projector outputs are not ordered. We can also use Hungarian algorithm~\cite{carion2020end} to find the correct matching between them and then facilitate the modeling of class-invariant feature. We will leave the application for images containing multiple classes such as object detection for future work.

\section{Discussion of WIDIn on Aligned Uni-Modal Models}

Training a joint \vlmm{} embedding space from scratch is computational expensive and how to utilize existing uni-modal model is important.
After all, as demonstrated in Table 1, when adapting a pre-trained model to a downstream tasks, though new knowledge are learned, training a single classifier (Linear Clf. and MLP CLf.) still results in knowledge forgetting. \approach{} is studied to mitigate this issue aligning uni-modal models from different modality.
At the same time, fusing information from two separately trained uni-modal models should also be of low cost while finetuning the entire network is still computational expensive. As such, for experiments in Sec. 4.2, only the feature space is updated. In this way, on the space that the \txtemb{s} reside, we learn visual representation that can be aligned with the language embedding and the whole space can be treated as a joint \vlmm{} embedding space. 

For the visual encoders listed in Table 2, the pre-training details are provided below. Meanwhile, we note all visual encoders listed below are only pre-trained on Imagenet, which is much less than the 400M data pairs used in CLIP training. 
\begin{itemize}[leftmargin=*]
    \item  Full Supervision is essentially pushing the image features towards the same class center, the features are clearly trained to be grouped according to the class label.
    \item DINO is trained by minimizing the difference of logits by the student encoder and by the teacher encoder in a self-supervised manner. The logits in DINO is obtained by projecting from the features through a linear layer and the projecter layer consisting of 65536 weight vectors. The weight vectors are jointly learned with the entire feature extractor. When we minimize the loss of DINO, for each image instance, we are pushing the features towards the same weight vector. As the number of instances in the train set is much larger than the number of weight vectors, the features from different instances with high similarity will be pushed towards the same class center. As such, the features learned from DINO can be properly clustered.
    \item MoCo is similar to SimCLR~\cite{chen2020simple} and directly compares the features. It is also studied in \cite{wang2020understanding} that contrastive learning aims to learning a uniform distribution across instances. 
    \item MAE is only trained to reconstructed the masked area and does not directly learns to matching instance features of high similarity.
\end{itemize}

\bitem{Version of language encoders}:
Admittedly, both CLIP-Text and BERT have different architectures in the published versions. For the fairness of comparison, we only choose the model version whose final the feature dimension is 768,
For the convenience of accommodation, we only select the model where a class token \texttt{<CLS>}~\cite{devlin2018bert} is learned.

\bitem{Implementation of Linear Clf. Baseline}.
For the fairness of comparison, Linear Clf. only train on the pre-extracted embeddings to learn a linear classifier.
However, for MAE and MoCo, we observed that it is essential to apply the data augmentation function during the training such that a more robust linear classifier can be learned. Furthermore, we found a batchnorm layer is added, \ie, the extracted \visemb{s} of the images first pass the batchnorm layer and the outputs of the batch norm layer are then fed into the classifier, in their official github repos.

\bitem{Learning Vision Encoder from Scratch}
In addition to use image encoder which is trained on an image-only dataset or a multi-modal dataset as intialization, we can also train an image encoder from scratch for domain generalization. 
However, as the training data is very limited, training the image encoder from scratch is very easy to overfit to the train set and the generalization of test data from the same domain is still poor.
In detail, on CUB-Painting, we train a ResNet18 from scratch via fully-supervised learning, the test accuracy (Src. 22.45, Tar. 14.59). Then, we test our approach after performing multi-modal contrastive learning between language embedding extracted by CLIP-Text and the image embedding to update the image encoder from scratch, and get the results (Src. 24.85, Tar. 15.30). Though it is still less acceptable, we can still see a slight gain by our approach.


\section{Experiment Implementation}

\bitem{Availability of Datasets}. All of the datasets used for experimental study in our approach are publicly available. 

\subsection{Training Details}

Thus, for fair comparison in Table 1, all approaches reported in Table 1 uses ViT-L$_{14}$ for experiments. In Table 2, we first choose ViT-B$_{16}$ for validation since public checkpoints on ViT-B$_{16}$ are provided by all approaches.
(ViT stands for Vision Transformer~\cite{dosovitskiy2020image} where ``L'' and ``B'', indicating the size of ViT, stand for large and base respectively. For ViT, each image is first split into non-overlapping grids and each grid is called a patch. Then, the number at sub-index means the size (\ie, height and width) of patch.)

During network training, we experimentally found that using SGD optimizer to minimize the alignment constraints based on contrastive learning is beneficial for the final performance. As such, even though the CLIP model is pre-trained by using AdamW optimizer, when we train the projector $\mathbf{F}_P$ by minimizing the loss $\mathcal{L}_{ia}+\mathcal{L}_{ca}$, we still use SGD as the optimizer with 0.002 as the learning rate. Then, we minimize the feature prediction loss $\mathcal{L}_{feat}$ and classification loss $\mathcal{L}_{cls}$, we use the AdamW as the optimizer with the learning rate 0.0001.

To balance the network optimization strength from different losses, we set the weight of $\mathcal{L}_{feat}$ as 2.0 while keep the weights for all other losses as 1.0. 
The training is very efficient and the maximum number of epochs we used for training is 60.
Meanwhile, the disentangler $F_D$ is implemented in a residual connection, \ie, we use $x+F_D(x)$ as the \cs{} visual embedding. In this way, from Eq. 3, $\visvec{}_e = \visvec{} - (\txtpseudo{} - \txtvec{}) = \visvec{}_e = \visvec{} + (\txtvec{} - \txtpseudo{})$. Then, following the definition of residual learning where $\visvec{}_e = \visvec{} + \mathbf{F}_D(\visvec{})$, the output of disentangler $\mathbf{F}_D(\visvec{})$ is in effect trained trained to mimic $\txtvec{} - \txtpseudo{}$.

For experiments on aligned uni-modal, we found using AdamW optimizer is more stable and we set the learning rate as 0.0001.

\subsection{Performance Details}

Both DomainNetMini and OfficeHome consist of one source domain and three targets domains during evaluation. In Table 1, we report the average of performance on three domains under Tar. and we below details the results on all three domains for reference DomainNetMini (95.99, 95.81, 96.21) and OfficeHome (90.64, 75.42, 96.00).

In Table 1, the approach ``Zero-Shot$^{*}$'' means adding domain description when extracting the \txtemb{} for zero-shot classification. For example, on DomainNetMini, we use ``the photo/image of an airplane'' for the approach ``Zero-Shot'' but use ``the clipart of an airplane'' for the approach ``Zero-Shot$^{*}$''. On Waterbirds, the domain is determined by the class and the environment. 

Then, for ablation study on three embedding types in Table 3(a), we note the discriminative information for classification may also be useful in indicating the domains, for example, the eyes of bird is natural images is smaller than that in painting/drawing. As such, the domain-specific discriminative information in \cs{} representation can also be useful for domain recognition. However, the experiments for object class recognition can still clearly demonstrate the effect of our approach.

\subsection{Comparison with other approaches}

\approach{} is different from conventional prompting algorithms which use prompting to learn embeddings for downstream task such as classification. In contrast, as explained in Line 223-233, \approach{} use prompting algorithm to learn \txtemb{} with fine-grained alignment, which is only used to disentangle the visual embedding but not directly used for downstream tasks.

By taking CoCoOp as one example, both CoCoOp and our \approach{} learn to map the visual embedding to word tokens and then extract embeddings from language encoder to facilitate the classification in downstream application. However, taking the sentence "The image of a dog" as an example, the token predicted by CoCoOp serves as prompt (e.g., "An image of a") and is concatenated with the class name. In contrast, \approach{} aims to map the visual embedding to a word of interest, e.g., "dog". According to the comparison below, the \approach{} achieves better performance and is also more efficient.

\begin{table}
\centering
\caption{Comparison with Conventional Prompting Approaches}
\resizebox{\linewidth}{!}
{
    \begin{tabular}{c|ccc|ccc}
    \hlineB{3}
    \multirow{2}{*}{Approach} & \multicolumn{3}{c|}{CUB-Painting} & \multicolumn{3}{c}{DomainNetMini} \\
                              & Src.      & Tar.      & Avg.     & Src.      & Tar.      & Avg.      \\ \hline
    Linear Clf.               & 85.91     & 64.33     & 75.12    & 95.03     & 93.75     & 94.39     \\
    CoOp                      & 83.34     & 65.41     & 74.38    & 96.12     & 95.46     & 95.79     \\
    CoCoOp                    & 86.3      & 67.91     & 77.11    & 96.81     & 95.88     & 96.35     \\
    WIDIn (Ours)               & 87.47     & 68.76     & 78.12    & 97.08     & 95.73     & 96.41    \\
    \hlineB{3}
    \end{tabular}
}
\label{tab:prompt}
\end{table}


\subsection{More Evaluation on CLIP embedding space}

Besides the experimental validation on CUB-Painting, DomainNetMini, Office-Home, where domain bias are defined by the class-agnostic environment (\eg, both dogs and cats may appear in the natural environment and part), we also evaluate our approach on Waterbirds~\cite{sagawadistributionally}, the domain bias is defined by the connection between class and environment.

\bitem{Waterbirds}~\cite{sagawadistributionally} consists of two classes, \ie, waterbirds and landbirds, but under different environments. During training, the environmental context is always matched with the class of birds (source), \ie, the landbirds are always on the land and the waterbirds are always on the water. However, during testing, there are images with mismatched environments (target), \ie, landbirds on the water and waterbirds on the land.
For evaluation purpose, we calculate the accuracy for each class at each environment. In this way, a domain is essentially determined by the class name and the environment. In other words, the model has only seen two domains at train time but four domains at test time. Ad such, we first detail the performance of zero-shot baseline and our approach below.

\begin{table}[]
    \centering
    \caption{Detailed results on WaterBirds.}
    \resizebox{0.8\linewidth}{!}
    {%
        \begin{tabular}{c|cc}
            \hlineB{3}
            Approach  & Zero-Shot & \approach{}   \\ \hline
            landbirds on land       & 99.45 & 99.31   \\
            waterbirds on water  & 97.64 & 98.70   \\
            landbirds on water       & 48.87 & 52.84   \\
            waterbirds on land  & 66.51 & 92.81   \\
            \hlineB{3}
        \end{tabular}
    }
\end{table}

\begin{table}[]
    \centering
    \caption{Comparison with Baseline.}
    \resizebox{0.7\linewidth}{!}
    {%
        \begin{tabular}{c|cc}
            \hlineB{3}
            Approach  & ~~~~Src.~~~~ & ~~~~Tar.~~~~   \\ \hline
            Zero-Shot       & 98.55 & 57.69   \\
            Zero-Shot*  & 98.55 & 58.77   \\
            Linear Clf.       & 92.75 & 72.80   \\
            \approach{}  & \textbf{99.00} & \textbf{72.83}   \\
            \hlineB{3}
        \end{tabular}\label{tab:wb_baseline}
    }
\end{table}

Since the classification task is still to differentiate waterbirds from landbirds, we choose to fuse the \txtemb{s} of descriptions of the same class, \eg, averaging the \txtemb{s} extracted from ``The photo of a landbird in the forest'' and ``The photo of a landbird on the water'' as the center of class ``landbird''. However, compared with Table~\ref{tab:wb_baseline}, the performance gain is not significant.

Further, as the class names are defined by the environment, we think the CLIP language embedding, e.g., extracted from ``The image of a landbird'', may naturally include the environment information of land (water) to the class name landbird (waterbird). As such, the effect of \cs{} visual embedding by \approach{} is not significantly shown comparing with visual embedding.
Thus, we think our \approach{} can still be further improved by using other description to represent these two classes, e.g., the description of attributes of the objects such as size and shape, which is not explicitly defined by the environment but can still be used to distinguish these two bird classes. Then, our framework can still be naturally generalized to these. We would like to leave it for future work.

\subsection{Ablation study of \approach{} on Domain Generalization}

\bitem{Training Schedule}. For our \approach{}, we first only learn the projector under the alignment constraint. Then, we learn the feature disentangler and classifier separately, \ie, the embedding $\visvec{}_e'$ predicted from $\mathbf{F}_D$ is not used to train $\mathbf{F}_C$. We represent the schedule as ($\mathbf{F}_P$, $\mathbf{F}_D$, $\mathbf{F}_C$).

\begin{table}[!htb]
    \centering
    \caption{Ablation on Training Schedule}
    \resizebox{.35\textwidth}{!}
    {%
        \begin{tabular}{cccc}
        \hlineB{3}
         Strategy  & Src.  & Tar.  & Avg.  \\ \hline
        ($\mathbf{F}_P$+$\mathbf{F}_C$, $\mathbf{F}_D$)      & 87.21 & 67.25 & 76.98 \\
        ($\mathbf{F}_P$+$\mathbf{F}_D$, $\mathbf{F}_C$)     & 86.37 & 64.82 & 75.59 \\
        ($\mathbf{F}_P$, $\mathbf{F}_D$+$\mathbf{F}_C$) & 86.76 & 66.10 & 76.43 \\ 
        \hlineB{3}
        \end{tabular}
    }
    \label{tab:schedule}
\end{table}

As shown in Table~\ref{tab:schedule}, in practice, we found that separating the training of the three modules result in the best performance. After all, at the early stage of the network training, the projector for predicting token embeddings has not been properly learned. Then, using the  $\visvec{}_e$ at early stage to guide the training of  predictor ($\mathbf{F}_P$+$\mathbf{F}_D$, $\mathbf{F}_C$) will confuse the network training and result in model collapsing.
Similarly, if the classifier is jointly trained with the predictor ($\mathbf{F}_P$+$\mathbf{F}_C$, $\mathbf{F}_D$), the model may show more bias to the source domain.
Meanwhile, if we use both $\visvec{}_e$ and $\visvec{}_e'$ to train the classifier ($\mathbf{F}_P$, $\mathbf{F}_D$+$\mathbf{F}_C$), the performance may drop slightly.
As such, we only use the $\visvec{}_e$ to train the classifier. In this way, as we are separating $\visvec{}_e$ and $\visvec{}_e'$ for loss calculation, we can learn the $F_C$ and $F_D$ at the same time. Meanwhile, when we train the $F_P$, we use the direction loss in LADS~\cite{dunlap2022using} as an alternative of MSE loss, and it underperforms our current design as Direction loss (CUB-Painting, \id: 86.81 and \ood:66.42) ignores the consistency of vector length.

\bitem{Sensitivity of scaler $k$}. Though we follow Eq. (3) with $k=1$ to roughly estimate the representation. As we normalized the embeddings in Eq.(3), the performance will be similar when $k$ is reasonable and WIDIn is robust to different selection of $k$. Specifically, we found the accuracy remains high when k $\leq$ 3 and only drops when
$k$ is too large to maintain stable training. Otherwise, the \ca{} feature maybe included. Meanwhile, we note the value of $k$ should also be adaptive for each instance. 


\end{document}